\documentclass[runningheads]{llncs}
\usepackage[T1]{fontenc}
\usepackage{graphicx}
\usepackage{booktabs}
\usepackage[misc]{ifsym}

\usepackage{amsmath}
\usepackage{amssymb}
\usepackage{cite}
\usepackage{url}
\usepackage[hidelinks]{hyperref}

\begin{document}

\title{Mirror-Fusion Attention for Reflection-Aware Self-Supervised Representation Learning}

\titlerunning{MFASSL}

\author{Ruixin Li\inst{1} \and
Jin Liu\inst{2} \and
Yuling Shi\inst{3} \and
Stefano Lodi~(\Letter)\inst{1}}

\authorrunning{R. Li et al.}

\institute{
Department of Computer Science and Engineering, University of Bologna, Bologna, Italy\\
\email{ruixin.li@studio.unibo.it, stefano.lodi@unibo.it}
\and
Jilin University, Changchun, China\\
\email{liujin0623@mails.jlu.edu.cn}
\and
Shanghai Jiao Tong University, Shanghai, China\\
\email{yuling.shi@sjtu.edu.cn}
}

\toctitle={Mirror-Fusion Attention for Reflection-Aware Self-Supervised Representation Learning}
\tocauthor={Ruixin Li, Jin Liu, Yuling Shi, Stefano Lodi}

\maketitle

\begin{center}
\small
Accepted at ECML PKDD 2026. The final authenticated version will be available in the Springer LNCS proceedings.
\end{center}

\begin{abstract}
Most self-supervised learning (SSL) methods encourage invariance across augmentations, but strict flip invariance can suppress informative left--right correspondences in approximately bilateral data such as medical images and human faces. We propose Mirror-Fusion-Augmented Self-Supervised Learning (MFASSL), a Vision Transformer framework that injects a soft reflection prior into standard SSL without redesigning the backbone. MFASSL constructs mirror-paired views aligned to an estimated symmetry axis and introduces a lightweight Mirror-Fusion Attention (MFA) module for adaptive token-level interaction between mirrored regions while preserving asymmetric cues. The base SSL objective is further coupled with reflection-consistency and mid-layer token-alignment losses. Across CheXpert, BraTS, CelebA-HQ, and WFLW, MFASSL improves downstream performance, calibration, and reflection robustness over MoCo-v3, DINO, and MAE baselines under matched ViT-B/16 settings. It also achieves stronger and more consistent gains than recent equivariant SSL approaches with only approximately 2.7\% additional parameters. These results show that lightweight geometry-aware priors can effectively complement invariance-based SSL. Code is publicly available at \url{https://github.com/Lirxstar/MFASSL}.


\end{abstract}

\section{Introduction}

Self-supervised learning (SSL) has become a central paradigm for visual representation learning, enabling large-scale pretraining without manual annotation. Existing methods broadly fall into two categories: discriminative frameworks, which align representations across augmented views through contrastive or self-distillation objectives~\cite{chen2020simple,chen2021empirical,NEURIPS2020_f3ada80d,caron2021emerging,zhou2021ibot}, and reconstruction-based frameworks, which learn local semantics by recovering masked content or features~\cite{he2022masked,wei2022masked,bao2021beit}. These paradigms have produced highly transferable representations across a wide range of domains, including recent medical foundation models~\cite{yao2025evax}. However, standard SSL pipelines generally do not treat reflection as meaningful structure: discriminative methods often suppress it through invariance, while reconstruction-based methods usually leave it implicit.

This limitation becomes particularly relevant in bilaterally organized data, where mirrored regions are structurally related but not strictly identical. In medical imaging, contralateral anatomy often shares broad spatial organization while differing in diagnostically important local detail; unilateral abnormalities in chest radiographs or asymmetric patterns in brain MRIs are therefore informative rather than nuisance variation~\cite{irvin2019chexpert}. A similar pattern arises in facial images, where mirrored regions remain correlated but can differ in expression, illumination, or localized attributes~\cite{wu2018look}. In such settings, treating reflection as a generic label-preserving transformation risks weakening precisely the asymmetric cues that downstream tasks rely on. A natural alternative is to move from strict invariance toward reflection-aware or equivariant representation learning, so that features respond predictably rather than being forced to suppress geometric structure. However, existing approaches often either impose symmetry through rigid architectural constraints or encourage it only through the training objective. For reflection-structured but imperfectly symmetric data, this leaves open the need for a lightweight representation-level mechanism that can model mirror correspondence while retaining informative asymmetry.

We address this gap with \emph{Mirror-Fusion-Augmented Self-Supervised Learning} (MFASSL), a simple framework for injecting a soft reflection prior into standard Vision Transformers. MFASSL forms mirror-paired views during pretraining and introduces a lightweight \emph{Mirror-Fusion Attention} (MFA) block that exchanges information between corresponding mirror tokens through adaptive gating. Rather than enforcing strict left--right equivalence, the model is trained to combine two complementary objectives: encouraging reflection-consistent structure where correspondence is reliable, and retaining local discrepancies where asymmetry is informative. This design keeps the backbone unchanged in spirit, while enabling reflection-aware reasoning at an intermediate representation level.

\section{Related Work}

\subsubsection{Equivariance and Symmetry Priors.}
Equivariance offers an alternative inductive bias to invariance by encouraging representations to transform predictably under geometric operations. Classical group-equivariant CNNs~\cite{cohen2016group,weiler2019general} and steerable architectures~\cite{cohen2016steerable} encode symmetry groups directly into the model, while SE(3)-equivariant networks and related transformer variants extend these ideas to richer geometric settings~\cite{fuchs2020se,xu20232,romero2020group,finzi2021practical,liao2022equiformer}. More recent work has introduced softer symmetry priors through self-supervised objectives or local consistency constraints. E-SSL~\cite{dangovski2021equivariant,zhang2022unsupervised} promotes transformation-aware features through equivariant learning objectives; EquiMod~\cite{devillers2023equimod} adds an equivariance module for visual instance discrimination; transformation-learning objectives directly train equivariant representations from self-supervised transformations~\cite{yu2024selfsupervised}; and OcticViT~\cite{nordstrom2025stronger} incorporates discrete symmetry groups into ViT-based SSL. Related pixel-level frameworks also encourage local geometric consistency through correspondence or uncertainty-guided alignment~\cite{wang2022uncertainty}. These studies demonstrate the usefulness of geometric priors for representation learning. However, their symmetry assumptions are usually introduced at the level of model design or optimization, rather than through an explicit feature-level interaction between mirror-paired regions. As a result, they may be less suitable for data with approximate bilateral regularity.

\subsubsection{Self-Supervised Learning in Medical Imaging.}
Self-supervised learning has become increasingly important in medical imaging~\cite{huang2023self,shurrab2022self,ma2022benchmarking}, where large volumes of unlabeled data are available but expert annotation is expensive. Early reconstruction-based approaches such as Models Genesis~\cite{zhou2019models} and TransVW~\cite{haghighi2021transferable} demonstrated the value of surrogate-task pretraining, and later contrastive, distillation-based, and masked-reconstruction methods further improved transfer across radiology and MRI applications~\cite{chaitanya2020contrastive,zhou2023self,moutakanni2024advancing}. More recently, medical foundation models have extended SSL to broader clinical imaging settings. Despite these advances, most medical SSL methods do not explicitly model contralateral or mirrored regions as structured correspondences. Some supervised or task-specific methods have incorporated bilateral or reflection priors for segmentation, adaptation, and brain-imaging analysis~\cite{han2021deep,yu2021bisenet,ma2024symmetryawareness}, but they typically rely on dense supervision, modality-specific assumptions, or downstream-task-specific formulations. As a result, reflection-aware SSL for general-purpose medical pretraining remains relatively underexplored.

\subsubsection{Symmetry and Reflection in Natural Vision.}
Beyond medical imaging, symmetry is also an important cue in natural vision and has been exploited in tasks such as facial analysis, human pose estimation, and fine-grained recognition. In facial image modeling, symmetry-aware methods use left--right consistency to improve completion or restoration of facial structure and appearance~\cite{zhang2018symmetry}. In pose estimation, mirror-based geometric constraints help reduce ambiguity in 3D human reconstruction~\cite{fang2021reconstructing}. In fine-grained recognition and canonical representation learning, symmetry can support semantically aligned templates and more structured feature spaces~\cite{lei2022cadex,joung2021learning}. Related two-branch designs in segmentation further show the benefit of structured feature interaction between global context and spatial detail~\cite{yu2018bisenet}. However, most of these methods use symmetry in task-specific or fixed-fusion ways rather than as a general prior for visual pretraining. More flexible handling of approximate reflection structure at the token level remains relatively underexplored.

\section{Method}
\label{sec:method}

\begin{figure*}[!t]
    \centering
    \includegraphics[width=\textwidth]{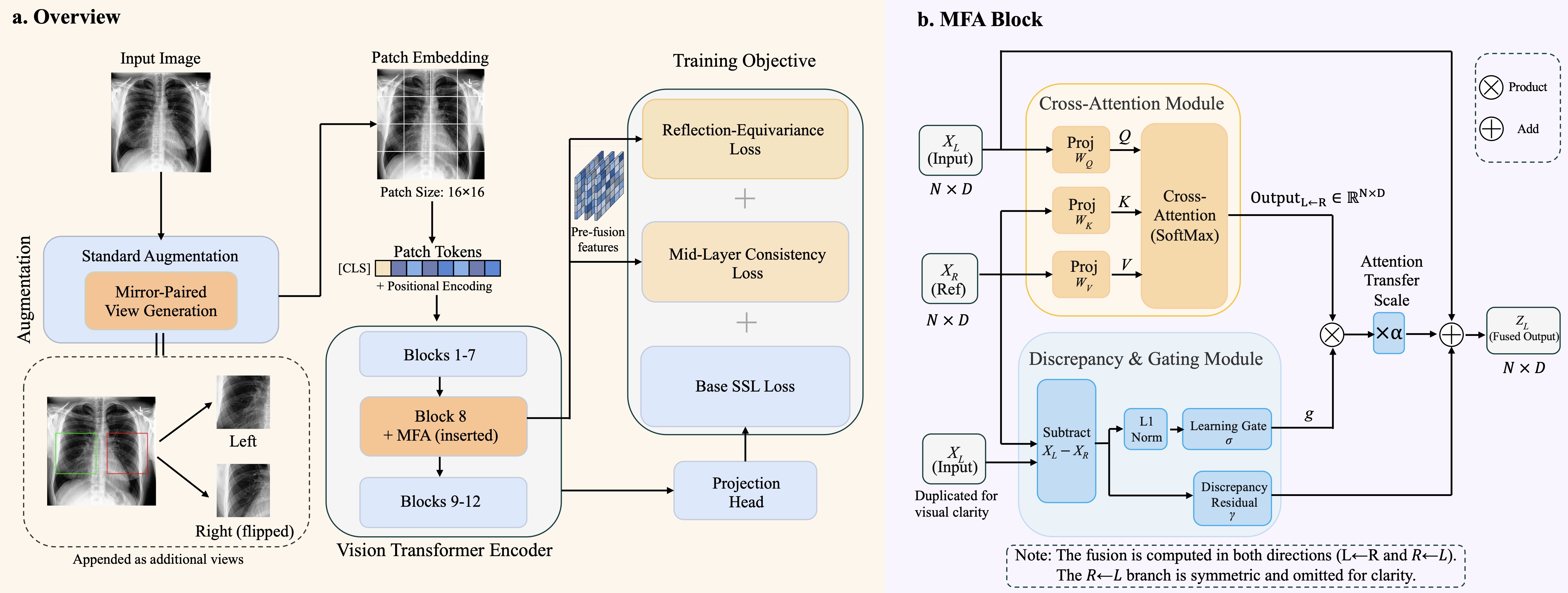}
    \caption{\textbf{MFASSL Architecture.} (\textbf{a}) Pretraining Pipeline: standard augmented views provide the base SSL supervision, while mirror-paired crops are routed through MFA during pretraining. Their pre-fusion tokens yield symmetry-aware losses ($\mathcal{L}_{\text{eq}}$, $\mathcal{L}_{\text{mid}}$) at layer 8, and their post-fusion representations are fed to the same base SSL objective. (\textbf{b}) Mirror-Fusion Attention (MFA): performs token-level fusion with a learnable gate and discrepancy residual, selectively preserving asymmetric structures. During training, the gated cross-view attention branch is gradually activated by a gate ramp, while the discrepancy residual is initialized conservatively through $\gamma$ so that its effect remains small early in training.}
    \label{fig:overview_mfa}
\end{figure*}

We propose Mirror-Fusion-Augmented Self-Supervised Learning (MFASSL), a plug-and-play framework for ViTs that injects a soft reflection prior into standard SSL. MFASSL combines mirror-paired view generation, a lightweight Mirror-Fusion Attention (MFA) block for token-level cross-mirror interaction, and a symmetry-aware objective for global and token-level reflection consistency. An overview of the MFASSL pretraining pipeline is shown in Fig.~\ref{fig:overview_mfa}(a).

A key design choice is that MFA is used as a pretraining-only adapter rather than as a replacement for a transformer block. Ordinary SSL views follow the standard ViT path, while mirror-paired views are processed to layer $\ell$, where pre-fusion tokens provide the inputs for the reflection-aware losses and MFA. The fused mirror-view representations then continue through the remaining transformer blocks and contribute to the same base SSL objective used by the underlying method. At inference time, MFA and mirror-paired inputs are removed, and the learned ViT is deployed with the standard single-image forward pass.

Pretraining data flow is as follows:
\begin{enumerate}
    \item Given an image $x$, a base SSL pipeline $B$, and a target layer $\ell$, sample ordinary SSL views $V=B(x)$ and encode them with the standard ViT path.
    \item Sample mirror crops $(x_L,x_R)$ around the jittered vertical axis; flip only $x_R$ for alignment.
    \item Run both crops to layer $\ell$ and store pre-fusion patch tokens $\phi_\ell(x_L),\phi_\ell(x_R)$.
    \item Compute $\mathcal{L}_{\text{eq}}$ and $\mathcal{L}_{\text{mid}}$ from these pre-fusion tokens.
    \item If $t<t_{\text{mfa}}$, continue with the unfused tokens; otherwise apply MFA and continue through blocks $\ell{+}1,\ldots,L$.
    \item Compute the base SSL objective using the ordinary SSL views together with the post-fusion mirror representations, then optimize Eq.~\ref{eq:loss-time}.
\end{enumerate}

\subsection{Mirror-Fusion Attention (MFA)}
\label{sec:mfa}
Given mirror-aligned token sequences $X_L, X_R \in \mathbb{R}^{N \times D}$ produced by a ViT encoder, MFA performs cross-view attention with learnable gating. Fusion is computed in both directions $L \leftarrow R$ and $R \leftarrow L$; for brevity, we describe the $L \leftarrow R$ branch only. MFA is a lightweight cross-attention block specialized for mirror-paired tokens and equipped with an additional discrepancy-preserving channel. The architecture is illustrated in Fig.~\ref{fig:overview_mfa}(b).

We use standard scaled dot-product cross-attention to model interactions between the original and mirrored features, with queries from $X_L$ and keys and values from $X_R$:
\begin{equation}
A_{L\leftarrow R} =
\mathrm{softmax}\!\left(
\frac{Q_L K_R^\top}{\sqrt{D_h}}
\right)V_R 
\in \mathbb{R}^{N \times D}.
\end{equation}
where $D_h$ denotes the attention feature dimension used for scaling.

\subsubsection{Learnable per-token distance-gating.}
To modulate fusion strength based on token-level spatial correspondence, we introduce a per-token gate. For the $i$-th token pair ($1 \leq i \leq N$):
\begin{equation}
g_i = \sigma\!\Big( a - b \, \textstyle\sum_{j=1}^{D}\sqrt{(X_{L,i,j} - X_{R,i,j})^2 + \epsilon^2} \Big),
\label{eq:gate}
\end{equation}
where $X_{L,i,j}$ and $X_{R,i,j}$ denote the $j$-th feature of the $i$-th token from $X_L$ and $X_R$ respectively, $a, b$ are learnable scalars, $\sigma$ is a sigmoid, and $\epsilon = 10^{-6}$. 

This computes a smooth $\ell_1$ distance over the feature dimension and yields a gate vector $g \in \mathbb{R}^{N \times 1}$, which is broadcast during fusion. The smoothed form is used instead of an exact $\ell_1$ norm to keep the gate differentiable at zero. The gate therefore allows MFA to suppress fusion for locally mismatched tokens while retaining stronger cross-view interaction at spatial positions with high mirror correspondence.

We enforce $b \geq 0$ via $b = \mathrm{softplus}(\tilde{b})$ and initialize $(b, a) = (1.0, 0.5)$, which biases the gate toward conservative fusion early in training. We note that the effective threshold depends on the feature magnitude at layer $\ell$; rather than a fixed geometric interpretation, this initialization is intended to keep the gate in a conservative, near-neutral regime for commonly observed mid-layer feature scales and to let it adapt during training.

\subsubsection{Fusion update with an asymmetry-preserving channel.}
The final fused representation combines identity, gated cross-view attention, and a discrepancy-preserving residual:
\begin{equation}
Z_L = X_L + \alpha \big( g \odot A_{L \leftarrow R} \big) + \gamma (X_L - X_R),
\label{eq:update}
\end{equation}
and symmetrically for $Z_R$, where $\alpha$ and $\gamma$ are learnable scalars (initialized conservatively to 0.1) and $\odot$ denotes element-wise multiplication. 

Structurally, the gated attention branch shares information across corresponding mirror locations, while the discrepancy term preserves non-symmetric evidence. When mirror correspondence is strong, the gated attention branch dominates cross-view interaction; when local asymmetry emerges, the discrepancy term helps retain informative differences. This design enables selective bilateral information exchange without enforcing strict left--right equivalence, while the small initialization improves training stability and mitigates early-stage distribution shift.

MFA is used only during pretraining. At inference time, the pretrained ViT encoder is applied normally to a single input image without mirror fusion. This choice is empirical rather than a formal invariance guarantee: the staged schedule and small residual initialization are used to keep the train--test feature shift small. To quantify this shift, we measure $\|Z_L-X_L\|/\|X_L\|$ at the fusion layer on the CheXpert validation set after training; the mean relative perturbation is 3.8\% ($\pm$1.2\%).

\subsection{Mirror-Paired View Generation}
\label{sec:mp-views}
Mirror-paired view construction is required in MFASSL because both the MFA gate and the token-level consistency loss assume spatial correspondence across the two views. A standard horizontal flip alone does not provide this: the two views remain in opposite orientations and therefore cannot be compared token-by-token without explicit alignment. To establish correspondence, we construct mirror-paired crops $(x_L, x_R)$ by slicing around a jittered vertical midline and horizontally flipping the right crop so that both views share the same spatial orientation, as illustrated in Fig.~\ref{fig:overview_mfa}(a). Throughout the paper, $x_R$ denotes this already aligned right crop, and no additional flip is applied in the losses. In our experiments, images are first resized and oriented using the dataset preprocessing protocol, and the symmetry axis is taken as the crop-center vertical line with a random horizontal jitter of up to 3\% of image width. Preliminary validation showed that 1\% jitter produced overly concentrated gates near the estimated midline, whereas 5\% jitter slightly degraded BraTS segmentation; we therefore keep 3\% fixed across all reported experiments. For domains with explicit landmarks or registration, the same construction can instead use the estimated anatomical or facial midline.

Mirroring is not applied to every augmentation. The standard SSL crops keep the original MoCo-v3, DINO, or MAE augmentation recipes, while the mirror pair is added as a separate paired branch. For multi-crop SSL methods such as MoCo-v3 and DINO, the mirror pair is appended as the last two crops in the multi-crop set, allowing reuse of the original training pipeline. For MAE, which does not use multi-crop training, $(x_L, x_R)$ is processed in a dedicated paired forward pass. Each crop is masked independently, MFA operates on the visible tokens at layer $\ell$, and the symmetry-aware losses $\mathcal{L}_{\text{eq}}$ and $\mathcal{L}_{\text{mid}}$ are computed only on token positions visible in both crops, while the reconstruction objective supplies the corresponding base SSL supervision after the paired branch continues through the decoder path.

\subsection{Symmetry-Aware Objective}
\label{sec:objective}
Let $\mathcal{L}_{\text{base}}$ denote the underlying SSL objective (contrastive, distillation-based, or reconstruction-based). MFASSL augments it with two reflection-aware terms:
\begin{equation}
\mathcal{L}_{\text{total}} = \mathcal{L}_{\text{base}} + \lambda_{\text{eq}} \mathcal{L}_{\text{eq}} + \lambda_{\text{mid}} \mathcal{L}_{\text{mid}}.
\label{eq:total-loss}
\end{equation}
Here, $\mathcal{L}_{\text{eq}}$ encourages global reflection-consistent alignment, while $\mathcal{L}_{\text{mid}}$ enforces token-level correspondence at the same pre-fusion layer where MFA operates. These two losses regularize the inputs to MFA. The base term is kept objective-compatible with the selected SSL framework: standard augmented views use the original MoCo-v3, DINO, or MAE objective, and the post-fusion mirror representations are passed through an identically formed branch of the same objective. Therefore, MFA parameters receive gradients through $\mathcal{L}_{\text{base}}$ without introducing an additional contrastive, distillation, or reconstruction loss family.

\subsubsection{Reflection-consistency loss.}
To enforce reflection-aware global consistency jointly with mid-layer consistency, we compute $\mathcal{L}_{\text{eq}}$ at the same designated layer as $\mathcal{L}_{\text{mid}}$, before fusion. We use a negative cosine similarity loss~\cite{chen2021exploring}:
\begin{equation}
\mathcal{L}_{\text{eq}} = 1 - \cos\big( s_\ell(x_L), s_\ell(x_R) \big),
\label{eq:leq}
\end{equation}
where $s_\ell(\cdot)$ denotes mean-pooled, $\ell_2$-normalized patch-token embeddings from layer $\ell$. Since the mirror crops are already spatially aligned (Sec.~\ref{sec:mp-views}), this loss encourages global bilateral feature similarity between corresponding left--right regions, promoting reflection-consistent representations at the designated layer. Cosine distance is used here because $s_\ell$ is a global representation and scale should not dominate the alignment signal.

\subsubsection{Mid-layer consistency loss.}
Head-level alignment alone does not adequately constrain token correspondences at the fusion layer. A token-wise constraint before MFA provides a stronger prior~\cite{romero2015fitnetshintsdeepnets} and stabilizes MFA once activated. Thus, we impose:
\begin{equation}
\mathcal{L}_{\text{mid}} = \frac{1}{N} \sum_{i=1}^{N} \Vert \hat{\phi}_\ell(x_L)_i - \hat{\phi}_\ell(x_R)_i \Vert_2^2,
\label{eq:lmid} 
\end{equation}
where $\hat{\phi}_\ell(\cdot)_i \in \mathbb{R}^D$ denotes the $\ell_2$-normalized token representation at spatial position~$i$ from transformer block $\ell$:
\begin{equation}
\hat{\phi}_{\ell,i} = \frac{\phi_{\ell,i}}{\|\phi_{\ell,i}\|_2},
\label{eq:phinorm}
\end{equation}
where $\phi_\ell \in \mathbb{R}^{N \times D}$ is the raw patch-token output of block $\ell$ and each token $\phi_{\ell,i} \in \mathbb{R}^D$ is independently normalized to unit $\ell_2$ norm.
Compared to prior pixel-level or region-level consistency losses~\cite{wang2022uncertainty,chaitanya2020contrastive}, $\mathcal{L}_{\text{mid}}$ is defined directly on mid-layer ViT tokens and is tightly aligned with the mirror-paired view construction in Sec.~\ref{sec:mp-views}. The squared $\ell_2$ form is applied after per-token normalization, so it acts as a token-wise cosine alignment up to a constant factor while preserving the FitNet-style token matching interpretation.

\subsection{Training and Implementation}
\label{sec:train}

We follow the base SSL optimizers and schedules commonly used for ViTs. To avoid noisy cross-branch coupling before mirror correspondence has formed, MFASSL uses a staged training strategy.

\subsubsection{(i) Symmetry-loss ramp}
For the first $T_{\text{sym}}$ epochs, we apply the symmetry-aware losses ($\mathcal{L}_{\text{eq}}, \mathcal{L}_{\text{mid}}$) with a linear ramp $w(t)$, allowing the backbone to learn stable, aligned representations from the symmetry prior before cross-mirror fusion:
\begin{equation}
w(t) = \mathrm{clip}\!\left( \frac{t}{T_{\text{sym}}}, 0, 1 \right),
\quad T_{\text{sym}} = 10 \ \text{epochs}.
\label{eq:sym-ramp}
\end{equation}
The total optimization objective at epoch $t$ is therefore
\begin{equation}
\mathcal{L}(t)
=
\mathcal{L}_{\text{base}}
+
w(t)\big[
\lambda_{\text{eq}} \mathcal{L}_{\text{eq}}
+
\lambda_{\text{mid}} \mathcal{L}_{\text{mid}}
\big].
\label{eq:loss-time}
\end{equation}

\subsubsection{(ii) MFA activation and gate ramp}
We disable MFA during the early stage of training and activate it only after the representation has become sufficiently aligned. Specifically, MFA is inserted at epoch $t_{\text{mfa}} = 12$, after which its gate is progressively released using
\begin{equation}
r_t = \mathrm{clip}\!\left(\frac{t - t_{\text{mfa}}}{T_{\text{gate}}}, 0, 1\right),
\qquad
g_t = r_t \, g,
\label{eq:gate-ramp}
\end{equation}
where $g \in \mathbb{R}^{N \times 1}$ denotes the gate vector defined by Eq.~\ref{eq:gate} for the current token sequence. With $T_{\text{gate}} = 10$ epochs, this schedule ensures that the perturbation introduced by gated cross-view fusion starts near zero and increases gradually. The discrepancy residual is not ramped directly. Instead, its early influence remains limited through the conservative initialization of $\gamma$, which keeps this branch small before stable left--right correspondence has formed.

For efficiency, standard crops are encoded first, and the mirror pair is processed once through a dedicated paired forward pass that returns both pre-fusion and post-fusion tokens. This keeps the additional computational cost modest relative to other equivariance-aware SSL formulations.

\section{Experiments}

\subsection{Datasets}
We use five datasets in total: CheXpert~\cite{irvin2019chexpert} and BraTS~\cite{li2024brain} as medical downstream benchmarks, OASIS-3~\cite{lamontagne2019oasis} as additional unlabeled MRI pretraining data, and CelebA-HQ~\cite{liu2018large} and WFLW~\cite{wu2018look} as natural-image downstream benchmarks. CheXpert contains 224{,}316 chest radiographs with 14 labels; for our multi-label classification experiments, we use frontal-view images. BraTS 2023 provides multi-modal brain MRI scans with annotations for three tumor subregions: enhancing tumor (ET), tumor core (TC), and whole tumor (WT). OASIS-3 provides additional unlabeled T1-weighted MRIs from an aging neuroimaging cohort. CelebA-HQ contains 30{,}000 high-resolution face images with 40 annotated attributes, and WFLW provides 98 facial landmarks per image. All medical splits are patient-wise.

\subsection{Pretraining Configuration}
\label{sec:pretraining}
All experiments use a ViT-B/16 backbone with 12 transformer blocks and a patch size of $16\times16$. All images are resized to $224\times224$. We evaluate three SSL paradigms: MoCo-v3, DINO, and MAE, and pretrain all models for 300 epochs using AdamW with cosine learning-rate decay.
MFASSL inserts the MFA block at block~8 and applies both the reflection-consistency loss $\mathcal{L}_{\text{eq}}$ and the mid-layer consistency loss $\mathcal{L}_{\text{mid}}$ at the same pre-fusion layer. After fusion, the mirror-pair branch continues through the remaining transformer blocks and is optimized with the original SSL objective, without changing the form of the base MoCo-v3, DINO, or MAE loss. We set $\lambda_{\text{eq}}=0.5$ and $\lambda_{\text{mid}}=1.0$, selected on the CheXpert validation set and then fixed for all other datasets and backbones. The symmetry-aware losses are linearly ramped up during the first 10 epochs, MFA is activated at epoch~12, and the gate is further released with a 10-epoch ramp. This gives a simple transfer heuristic: use a middle-late layer (about two-thirds depth), keep the two symmetry-loss weights fixed, and only reduce the weights if the base SSL loss becomes unstable. Appendix summarizes the consolidated hyperparameter settings.

For medical classification, pretraining is performed on CheXpert. For MRI-based segmentation and robustness experiments, we jointly pretrain on unlabeled BraTS and OASIS-3 using 50/50 mixed mini-batches. For natural-image experiments, pretraining is performed on CelebA-HQ. At inference time, MFA is removed and the finetuned model remains a standard ViT encoder.
\subsubsection{Controlled baselines.}
To isolate the effect of the proposed reflection-aware components, all baselines are trained from random initialization without ImageNet pretraining, dataset-specific augmentation tuning, or test-time augmentation. This standardized setting enables a cleaner assessment of method-level differences under matched training conditions, but it also limits direct comparison with ImageNet-initialized pipelines and flip test-time ensembling. We therefore interpret MFASSL as improving the single-forward-pass encoder under matched pretraining rather than as a replacement for all inference-time augmentation strategies.
\subsubsection{Competitor methods.}
We compare against two recent equivariant SSL approaches under the same ViT-B/16 backbone and training budget:
(i)~\textbf{E-SSL}~\cite{dangovski2021equivariant}, which adds an equivariant prediction loss to the base SSL objective, and
(ii)~\textbf{OcticViT}~\cite{nordstrom2025stronger}, which embeds discrete symmetry groups into the ViT architecture. We evaluate two OcticViT variants, OcticViT-H$_8$ and OcticViT-I$_8$, using the hyperoctahedral and icosahedral groups of order~8, respectively. OcticViT is evaluated only under the DINO backbone because its group-equivariant architectural modifications, specifically the group-structured patch token projections, are incompatible with MoCo-v3's momentum contrast objective and MAE's masked-token reconstruction pipeline. E-SSL is applied to DINO, MoCo-v3 and MAE. All competitors are re-implemented and trained under identical settings.

\subsection{Evaluation Protocols}
CheXpert is evaluated with both linear probing and full fine-tuning. BraTS segmentation is evaluated slice-wise in 2D, reporting Dice and Hausdorff distance at the 95th percentile (HD95) for the three standard subregions (ET, TC, WT) and their means, as commonly done in nnU-Net~\cite{isensee2021nnu}, together with voxel-level calibration metrics: expected calibration error (ECE~\cite{guo2017calibration}) and negative log-likelihood (NLL~\cite{minderer2021revisiting}). In the main text, we report mean Dice and mean HD95 for compact comparison, while the full ET/TC/WT breakdown is provided in the Appendix. CelebA-HQ is evaluated using classification accuracy, NLL, ECE, and Flip-Consistency. WFLW is evaluated using normalized mean error (NME), AUC@0.1, Failure@0.1, and Flip-Consistency. For BraTS we report mean $\pm$ standard deviation over three runs. For the other benchmarks, the main tables report point estimates averaged over repeated runs.

Flip-Consistency measures prediction agreement between an image and its horizontal reflection. For CheXpert, it is the fraction of test images for which the predicted multi-label set is unchanged after reflection. For CelebA-HQ, it is computed over all 40 attribute predictions. For WFLW, flipped predictions are first remapped to the corresponding landmark indices before consistency is measured. 

\subsection{Main Results}
\label{sec:main-results}
Across four representative downstream domains, CheXpert, BraTS, CelebA-HQ, and WFLW, MFASSL generally improves task performance and reflection-related consistency under identical ViT-B/16 backbones and training budgets (Tables~\ref{tab:chexpert}--\ref{tab:celeba_wflw}). The gains are strongest in fine-tuning and remain visible across contrastive, distillation-based, and reconstruction-based SSL backbones. Linear-probe gains are smaller and mixed, so we do not claim that MFASSL universally improves frozen representations; rather, the evidence supports better fine-tunable initialization and reflection-aware robustness.

\begin{table*}[t]
\centering
\caption{\textbf{CheXpert (14 labels).} Linear-probe and fine-tuning results under the same ViT-B/16 backbone and training budget. Flip denotes Flip-Consistency, which measures prediction agreement under horizontal reflection. Competitor methods are evaluated under matched settings. Lower is better for NLL, ECE, and Brier.}
\label{tab:chexpert}
\begin{tabular}{llccccccc}
\toprule
Model & Probe & AUROC$\uparrow$ & AUPRC$\uparrow$ & F1$\uparrow$ & NLL$\downarrow$ & ECE$\downarrow$ & Brier$\downarrow$ & Flip(\%)$\uparrow$ \\
\midrule
DINO & Linear & 80.12 & 68.40 & 55.72 & 0.442 & 0.036 & 0.091 & 89.2 \\
\quad + E-SSL & Linear & 80.20 & 68.70 & 55.85 & 0.439 & 0.036 & 0.091 & 89.1 \\
\quad + OcticViT-H$_8$ & Linear & 80.15 & 68.50 & 55.71 & 0.441 & 0.037 & 0.089 & 88.8 \\
\quad + OcticViT-I$_8$ & Linear & 80.06 & 68.31 & 55.63 & 0.443 & 0.038 & 0.091 & 89.2 \\
\quad ours & Linear & \textbf{80.35} & \textbf{69.10} & \textbf{56.08} & \textbf{0.431} & 0.037 & 0.090 & 89.0 \\
\cmidrule{1-9}
MoCo-v3 & Linear & 78.94 & 65.83 & \textbf{54.28} & 0.461 & 0.041 & 0.095 & 87.3 \\
\quad + E-SSL & Linear & 79.18 & 66.20 & 54.00 & 0.458 & 0.041 & 0.095 & \textbf{87.4} \\
\quad ours & Linear & \textbf{79.51} & \textbf{66.60} & 54.06 & \textbf{0.454} & \textbf{0.040} & \textbf{0.094} & 87.2 \\
\cmidrule{1-9}
MAE & Linear & 77.86 & 63.47 & 52.93 & 0.486 & 0.044 & 0.100 & 88.2 \\
\quad + E-SSL & Linear & 78.30 & 64.05 & 53.05 & 0.480 & 0.044 & 0.099 & 88.3 \\
\quad ours & Linear & \textbf{78.92} & \textbf{64.49} & \textbf{53.20} & \textbf{0.474} & \textbf{0.043} & \textbf{0.098} & \textbf{88.5} \\
\midrule
DINO & Finetune & 84.72 & 74.88 & 60.42 & 0.372 & 0.039 & 0.079 & 90.5 \\
\quad + E-SSL & Finetune & 84.88 & 74.95 & 60.65 & 0.370 & 0.038 & 0.079 & 90.7 \\
\quad + OcticViT-H$_8$ & Finetune & 84.81 & 74.92 & 60.62 & 0.371 & 0.036 & 0.079 & 90.3 \\
\quad + OcticViT-I$_8$ & Finetune & 84.95 & 74.96 & 60.71 & 0.367 & 0.037 & 0.078 & 90.9 \\
\quad ours & Finetune & \textbf{85.96} & \textbf{76.05} & \textbf{61.88} & \textbf{0.360} & \textbf{0.029} & \textbf{0.077} & \textbf{92.4} \\
\cmidrule{1-9}
MoCo-v3 & Finetune & 83.94 & 73.25 & 59.88 & 0.386 & \textbf{0.031} & 0.082 & 90.1 \\
\quad + E-SSL & Finetune & 84.20 & 73.60 & 59.95 & 0.383 & \textbf{0.031} & 0.082 & 90.2 \\
\quad ours  & Finetune & \textbf{84.50} & \textbf{73.98} & \textbf{60.10} & \textbf{0.373} & 0.032 & \textbf{0.081} & \textbf{91.0} \\
\cmidrule{1-9}
MAE & Finetune & 81.62 & 70.94 & 57.91 & 0.407 & \textbf{0.037} & 0.089 & 89.1 \\
\quad + E-SSL & Finetune & 81.95 & 71.30 & 58.05 & 0.404 & \textbf{0.037} & 0.089 & 89.3 \\
\quad ours  & Finetune & \textbf{82.31} & \textbf{71.85} & \textbf{58.22} & \textbf{0.396} & 0.038 & \textbf{0.088} & \textbf{90.6} \\
\bottomrule
\end{tabular}
\end{table*}

\begin{table}[t]
\centering
\caption{\textbf{BraTS 2023 segmentation (2D slice-wise).} Mean $\pm$ std over three runs. Lower is better for HD95/ECE/NLL.}
\label{tab:brats}
\begin{tabular}{lcccc}
\toprule
\textbf{Model} & \textbf{Dice}$\uparrow$ & \textbf{HD95 (mm)}$\downarrow$ & \textbf{ECE}$\downarrow$ & \textbf{NLL}$\downarrow$ \\
\midrule
DINO & 0.827 $\pm$ 0.016 & 8.6 $\pm$ 1.2 & 0.047 $\pm$ 0.010 & 0.226 $\pm$ 0.014 \\
\quad + E-SSL & 0.830 $\pm$ 0.015 & 8.4 $\pm$ 1.1 & 0.046 $\pm$ 0.010 & 0.224 $\pm$ 0.014 \\
\quad + OcticViT-H$_8$ & 0.829 $\pm$ 0.015 & 8.5 $\pm$ 1.1 & 0.046 $\pm$ 0.010 & 0.225 $\pm$ 0.014 \\
\quad + OcticViT-I$_8$ & 0.828 $\pm$ 0.015 & 8.3 $\pm$ 1.1 & 0.046 $\pm$ 0.009 & 0.221 $\pm$ 0.013 \\
\quad ours & \textbf{0.836 $\pm$ 0.014} & \textbf{8.1 $\pm$ 1.0} & \textbf{0.044 $\pm$ 0.009} & \textbf{0.214 $\pm$ 0.012} \\
\midrule
MoCo-v3 & 0.817 $\pm$ 0.018 & 9.4 $\pm$ 1.3 & 0.052 $\pm$ 0.010 & 0.244 $\pm$ 0.015 \\
\quad + E-SSL & 0.820 $\pm$ 0.018 & 9.1 $\pm$ 1.3 & 0.052 $\pm$ 0.010 & 0.241 $\pm$ 0.015 \\
\quad ours & \textbf{0.825 $\pm$ 0.016} & \textbf{8.9 $\pm$ 1.1} & \textbf{0.048 $\pm$ 0.010} & \textbf{0.229 $\pm$ 0.013} \\
\midrule
MAE & 0.843 $\pm$ 0.013 & 7.9 $\pm$ 1.0 & \textbf{0.041 $\pm$ 0.009} & 0.206 $\pm$ 0.012 \\
\quad + E-SSL & 0.846 $\pm$ 0.013 & 7.9 $\pm$ 1.0 & \textbf{0.041 $\pm$ 0.009} & 0.204 $\pm$ 0.012 \\
\quad ours & \textbf{0.851 $\pm$ 0.011} & \textbf{7.5 $\pm$ 0.9} & 0.043 $\pm$ 0.008 & \textbf{0.190 $\pm$ 0.011} \\
\bottomrule
\end{tabular}
\end{table}

\begin{table}[t]
\centering
\caption{\textbf{Attribute classification and landmark localization benchmarks.}
Abbreviations: Acc. = accuracy, Flip = Flip-Consistency, AUC = AUC@0.1, and Fail = Failure@0.1. CelebA-HQ Acc./Flip and WFLW NME/Fail are reported in \%. Higher is better for Acc., Flip, and AUC; lower is better for NLL, ECE, NME, and Fail.}
\label{tab:celeba_wflw}
\begin{tabular}{lcccc|cccc}
\toprule
& \multicolumn{4}{c|}{\textbf{CelebA-HQ}} & \multicolumn{4}{c}{\textbf{WFLW (98 landmarks)}} \\
\cmidrule(lr){2-5} \cmidrule(lr){6-9}
\textbf{Model}
& \textbf{Acc.} $\uparrow$
& \textbf{Flip} $\uparrow$
& \textbf{NLL} $\downarrow$
& \textbf{ECE} $\downarrow$
& \textbf{NME} $\downarrow$
& \textbf{AUC} $\uparrow$
& \textbf{Fail} $\downarrow$
& \textbf{Flip} $\uparrow$ \\
\midrule
DINO & 90.3 & 91.5 & 0.195 & 0.031 & 4.61 & 0.528 & 3.4 & 0.924 \\
DINO (ours) & \textbf{91.2} & \textbf{93.6} & \textbf{0.171} & \textbf{0.024} & 4.46 & 0.542 & 3.0 & 0.938 \\
MoCo-v3 & 89.4 & 90.6 & 0.214 & 0.035 & 4.74 & 0.517 & 3.7 & 0.919 \\
MoCo-v3 (ours) & \textbf{90.6} & \textbf{92.8} & \textbf{0.186} & \textbf{0.027} & 4.58 & 0.532 & 3.3 & 0.933 \\
MAE & 90.7 & 92.1 & 0.182 & 0.027 & 4.55 & 0.536 & 3.2 & 0.925 \\
MAE (ours) & \textbf{91.1} & \textbf{94.0} & \textbf{0.160} & \textbf{0.021} & \textbf{4.39} & \textbf{0.551} & \textbf{2.9} & \textbf{0.942} \\
\bottomrule
\end{tabular}
\end{table}

\subsubsection{Medical benchmarks: CheXpert and BraTS.}
On CheXpert (Table~\ref{tab:chexpert}), MFASSL delivers the strongest overall improvements across all three SSL baselines. Under fine-tuning, it raises AUROC from 84.72 to 85.96 on DINO, from 83.94 to 84.50 on MoCo-v3, and from 81.62 to 82.31 on MAE, while also improving AUPRC, F1, and Flip-Consistency in each case. Competing symmetry-aware baselines yield smaller gains: for example, on DINO, E-SSL improves AUROC by only +0.16\,pp and OcticViT by +0.23\,pp, compared with +1.24\,pp for MFASSL. Calibration is also generally improved, especially in fine-tuning, with DINO ECE decreasing from 0.039 to 0.029 and NLL from 0.372 to 0.360. The linear-probe block is more modest: DINO and MAE improve slightly, while MoCo-v3 F1 decreases from 54.28 to 54.06 despite AUROC and AUPRC gains. We therefore treat the linear-probe result as evidence that the frozen representation is largely preserved, not as the main source of the method's benefit. On BraTS (Table~\ref{tab:brats}), MFASSL similarly improves segmentation quality across all three backbones, achieving the best mean Dice on DINO (0.836), MoCo-v3 (0.825), and MAE (0.851), together with lower HD95 in each setting. The largest gain appears on DINO, where mean Dice increases from 0.827 to 0.836 and HD95 decreases from 8.6\,mm to 8.1\,mm, whereas OcticViT variants provide only marginal Dice gains under the same protocol. MAE + MFASSL shows a small ECE increase (0.041 $\rightarrow$ 0.043), but still attains the best overall Dice and NLL (0.190). Overall, the medical results in Tables~\ref{tab:chexpert} and~\ref{tab:brats} show that MFASSL improves predictive quality and reflection consistency in the evaluated settings, with calibration gains that are strongest for DINO and MoCo-v3.

\subsubsection{Natural-image benchmarks: CelebA-HQ and WFLW.}
The same trend extends beyond medical data. On CelebA-HQ and WFLW (Table~\ref{tab:celeba_wflw}), MFASSL improves performance across all three SSL backbones. For CelebA-HQ, accuracy increases from 90.3 to 91.2 on DINO, from 89.4 to 90.6 on MoCo-v3, and from 90.7 to 91.1 on MAE; Flip-Consistency also rises consistently, reaching 93.6, 92.8, and 94.0, respectively. For WFLW, MFASSL reduces NME from 4.61 to 4.46 on DINO, from 4.74 to 4.58 on MoCo-v3, and from 4.55 to 4.39 on MAE, while also improving Fail@0.1 and Flip consistency. The best overall WFLW result is obtained by MAE + MFASSL, which reaches 4.39 NME and 2.9 Fail@0.1 (Table~\ref{tab:celeba_wflw}). 

\subsection{Ablation and Extended Analysis}
\label{sec:ablation}
We conduct controlled ablations on CheXpert and BraTS using the DINO backbone to isolate the effects of mirrored inputs, symmetry-aware losses, and MFA, and then examine the impact of insertion depth and backbone scale.

\subsubsection{Component ablation.}
Table~\ref{tab:combined_ablation} shows that naive mirrored-input duplication has negligible effect, changing CheXpert AUROC from 84.72 to 84.73 and leaving BraTS mean Dice unchanged at 0.827. Adding the symmetry-aware losses alone already improves performance on both datasets, raising AUROC to 84.99 and Dice to 0.831. By contrast, MFA without $\mathcal{L}_{\text{eq}}$ or $\mathcal{L}_{\text{mid}}$ yields smaller gains than the full formulation, indicating that its benefit is realized most clearly when paired with symmetry-aware supervision. The full model performs best, reaching 85.96 AUROC on CheXpert and 0.836 mean Dice with 8.1\,mm HD95 on BraTS. Together, these results show that the main gains come from combining feature-level mirror interaction with explicit symmetry-aware supervision, rather than from mirrored views or added module capacity alone.

\begin{table*}[t]
\centering
\caption{\textbf{Ablation study on CheXpert and BraTS using the DINO backbone.} We isolate the contribution of mirrored inputs, symmetry-aware losses, and MFA. The "Mirrored input only'' row augments pretraining with the mirrored image view only, without using $\mathcal{L}_{\text{eq}}$, $\mathcal{L}_{\text{mid}}$, or MFA; therefore the three component columns remain marked as absent. For BraTS, mDice and mHD95 denote mean Dice and mean HD95 across ET/TC/WT. Lower is better for ECE and mHD95. The full metric version is provided in Appendix.}

\label{tab:combined_ablation}
\begin{tabular}{lccc|cc|cc}
\toprule
& & & & \multicolumn{2}{c|}{\textbf{CheXpert}} & \multicolumn{2}{c}{\textbf{BraTS}} \\
\cmidrule(lr){5-6} \cmidrule(lr){7-8}
\textbf{Model Variant} & $\mathcal{L}_{\text{eq}}$ & $\mathcal{L}_{\text{mid}}$ & \textbf{MFA} &
\textbf{AUROC}$\uparrow$ & \textbf{ECE}$\downarrow$ &
\textbf{mDice}$\uparrow$ & \textbf{mHD95}$\downarrow$ \\
\midrule
Baseline & -- & -- & -- & 84.72 & 0.039 & 0.827 & 8.6 \\
Mirrored input only & -- & -- & -- & 84.73 & 0.038 & 0.827 & 8.6 \\
MFA (only) & -- & -- & \checkmark & 84.86 & 0.037 & 0.828 & 8.5 \\
Mirrored + $\mathcal{L}_{\text{eq}}$ & \checkmark & -- & -- & 84.71 & 0.037 & 0.829 & 8.5 \\
Mirrored + $\mathcal{L}_{\text{eq}}+\mathcal{L}_{\text{mid}}$ & \checkmark & \checkmark & -- & 84.99 & 0.035 & 0.831 & 8.4 \\
Full (ours) & \checkmark & \checkmark & \checkmark & \textbf{85.96} & \textbf{0.029} & \textbf{0.836} & \textbf{8.1} \\
\bottomrule
\end{tabular}
\end{table*}

\subsubsection{Layer placement.}
Table~\ref{tab:layer} evaluates representative insertion depths for MFA and the symmetry-aware losses in the 12-layer ViT-B/16 backbone. Layers~4 and~6 diverge during training, suggesting that early features are too unstable for reliable mirror correspondence. In these failed runs, instability appears shortly after MFA activation: large pre-fusion token discrepancies lead to low-entropy gate values and large paired-branch gradients, causing the token-alignment term to dominate early patch features. Among the convergent settings, layer~8 gives the best performance, while layers~10 and~12 are slightly weaker. This pattern indicates that mid-level representations provide the best balance between spatial structure and semantic maturity for reflection-aware fusion. In practice, we monitor the MFA gate distribution and the gradient norm of the paired branch; if either becomes unstable, delaying $t_{\mathrm{mfa}}$ or reducing $\lambda_{\mathrm{mid}}$ is safer than moving MFA to an earlier layer.

\begin{table}[t]
\centering
\caption{\textbf{Layer placement ablation (CheXpert, DINO(ours)).} MFA and both symmetry losses are placed at the indicated layer. Layers 4 and 6 cause gradient explosion.}
\label{tab:layer}
\begin{tabular}{lcccccc}
\toprule
\textbf{Layer $\ell$} & \textbf{Status} & \textbf{AUROC}$\uparrow$ & \textbf{AUPRC}$\uparrow$ & \textbf{F1}$\uparrow$ & \textbf{ECE}$\downarrow$ & \textbf{NLL}$\downarrow$ \\
\midrule
4  & Diverged & --- & --- & --- & --- & --- \\
6  & Diverged & --- & --- & --- & --- & --- \\
\textbf{8} & \textbf{Converged} & \textbf{85.96} & \textbf{76.05} & \textbf{61.88} & \textbf{0.029} & \textbf{0.360} \\
10 & Converged & 85.63 & 75.77 & 61.60 & 0.031 & 0.363 \\
12 & Converged & 85.41 & 75.58 & 61.39 & 0.032 & 0.364 \\
\bottomrule
\end{tabular}
\end{table}

\subsubsection{Multi-Architecture Evaluation}
\label{sec:arch-ablation}
Table~\ref{tab:arch} shows that the gains are not specific to ViT-B/16. On ViT-S/16, MFASSL consistently improves CheXpert AUROC by 0.85--0.88\,pp across DINO, MoCo-v3, and MAE, and improves BraTS mean Dice by 0.7--0.8\,pp. The ViT-B/16 results follow the same trend, with AUROC gains of 0.56--1.24\,pp and Dice gains of 0.8--0.9\,pp. Overall, MFASSL remains effective across both backbone scales and all three SSL paradigms.

\begin{table}[t]
\centering
\caption{\textbf{Multi-architecture evaluation.} CheXpert (finetune) and BraTS (segmentation) results for ViT-S/16 and ViT-B/16 across all three SSL paradigms. Lower is better for ECE, HD95, and NLL.}
\label{tab:arch}
\begin{tabular}{llccccccc}
\toprule
& & \multicolumn{4}{c}{\textbf{CheXpert (Finetune)}} & \multicolumn{3}{c}{\textbf{BraTS}} \\
\cmidrule(lr){3-6} \cmidrule(lr){7-9}
\textbf{Backbone} & \textbf{Method} & AUROC$\uparrow$ & AUPRC$\uparrow$ & F1$\uparrow$ & ECE$\downarrow$ & Dice$\uparrow$ & HD95$\downarrow$ & NLL$\downarrow$ \\
\midrule
ViT-S/16 & DINO & 83.64 & 73.52 & 59.38 & 0.043 & 0.824 & 8.8 & 0.235 \\
         & DINO(ours) & \textbf{84.52} & \textbf{74.44} & \textbf{60.58} & \textbf{0.035} & \textbf{0.831} & \textbf{8.4} & \textbf{0.224} \\
ViT-S/16 & MoCo-v3 & 83.31 & 73.18 & 59.10 & 0.044 & 0.816 & 9.5 & 0.246 \\
         & MoCo-v3(ours) & \textbf{84.18}  & \textbf{73.96} & \textbf{60.24} & \textbf{0.036} & \textbf{0.824} & \textbf{9.0} & \textbf{0.233} \\
ViT-S/16 & MAE & 83.45 & 73.29 & 59.22 & 0.044 & 0.823 & 8.9 & 0.238 \\
         & MAE(ours) & \textbf{84.30} & \textbf{74.10} & \textbf{60.36} & \textbf{0.036} & \textbf{0.830} & \textbf{8.5} & \textbf{0.227} \\
\midrule
ViT-B/16 & DINO & 84.72 & 74.88 & 60.42 & 0.039 & 0.827 & 8.6 & 0.226 \\
         & DINO(ours) & \textbf{85.96} & \textbf{76.05} & \textbf{61.88} & \textbf{0.029} & \textbf{0.836} & \textbf{8.1} & \textbf{0.214} \\
ViT-B/16 & MoCo-v3 & 83.94 & 73.25 & 59.88 & \textbf{0.031} & 0.817 & 9.4 & 0.244 \\
         & MoCo-v3(ours) & \textbf{84.50} & \textbf{73.98} & \textbf{60.10} & 0.032 & \textbf{0.825} & \textbf{8.9} & \textbf{0.229} \\
ViT-B/16 & MAE & 81.62 & 70.94 & 57.91 & \textbf{0.037} & 0.843 & 7.9 & 0.206 \\
         & MAE(ours)  & \textbf{82.31} & \textbf{71.85} & \textbf{58.22} & 0.038 & \textbf{0.851} & \textbf{7.5} & \textbf{0.190} \\
\bottomrule
\end{tabular}
\end{table}

\section{Discussion}
\label{sec:discussion}

\subsubsection{Limitations.} MFASSL has several limitations that define its current scope. First, the method assumes an approximately known vertical reflection axis, so its effectiveness may decrease when this axis is poorly aligned, ambiguous, or semantically weak. Our current jitter handles small horizontal shifts but does not fully evaluate angular midline errors such as $\pm5^\circ$ or $\pm10^\circ$ rotations. Second, MFASSL is intended for bilaterally structured domains; on scenes, cluttered objects, or other weakly symmetric images, the mirror prior may become uninformative or harmful, and such settings require explicit validation rather than direct extrapolation. Third, the current formulation is restricted to planar bilateral symmetry and does not yet cover richer structural priors such as rotational symmetries, multiple axes, or learned symmetry groups. Fourth, the most effective MFA insertion depth may vary across backbone families and scales, and early-layer insertion remains unstable in our current implementation. Finally, our study uses matched pretraining from scratch and single-forward-pass inference; ImageNet-initialized pretraining and flip test-time averaging are complementary baselines that should be compared in larger-scale follow-up studies. The residual inference protocol should be regarded as an empirical property of the present design rather than a general guarantee for stronger fusion variants.

\subsubsection{Conclusion.} We presented Mirror-Fusion-Augmented Self-Supervised Learning (MFASSL), a reflection-aware training framework that augments standard SSL with mirror-paired supervision and selective mid-layer interaction, without requiring a redesigned backbone. Across the evaluated settings, the results indicate that soft symmetry guidance can complement invariance-driven self-supervision and improve representation quality when bilateral structure is informative. These findings suggest that lightweight, geometry-aware training priors are a promising direction for future self-supervised representation learning in vision.

 \bibliographystyle{splncs04}
 \bibliography{mybibliography}

\clearpage
\appendix
\section*{\centering Appendix}
\addcontentsline{toc}{section}{Appendix}
\setcounter{table}{0}
\renewcommand{\thetable}{A\arabic{table}}
\renewcommand{\theHtable}{appendix.table.\arabic{table}}
\renewcommand{\theHsection}{appendix.section.\arabic{section}}
\renewcommand{\theHsubsection}{appendix.subsection.\arabic{section}.\arabic{subsection}}

\raggedbottom

\section{Relation of MFA to Reflection Equivariance}
\label{app:theory}

We briefly relate MFA to equivariance theory in order to clarify its design objective. 
Let $\mathcal{R}$ denote horizontal reflection, and let $f_\theta : \mathcal{X} \rightarrow \mathcal{Z}$ be an encoder. 
Standard invariance-based self-supervised learning encourages
\begin{equation}
f_\theta(\mathcal{R}x) = f_\theta(x),
\label{eq:app_inv}
\end{equation}
which removes reflection-related variation by mapping an image and its reflected version to the same representation. 
By contrast, strict reflection equivariance requires
\begin{equation}
f_\theta(\mathcal{R}x) = \rho(\mathcal{R}) f_\theta(x),
\label{eq:app_eq}
\end{equation}
where $\rho(\mathcal{R})$ denotes the induced action in feature space. 
This preserves reflection structure by requiring the representation to transform predictably under reflection.

However, real bilaterally structured data are usually only approximately symmetric. 
In chest radiographs, brain MR images, and faces, global bilateral regularity often coexists with meaningful local asymmetry. 
As a result, strict invariance may suppress useful left--right differences, whereas exact equivariance everywhere may be unnecessarily rigid. 
MFASSL is designed for this intermediate regime: rather than enforcing a hard group-equivariant constraint, it introduces a soft reflection-aware bias that encourages reflection-consistent processing where bilateral correspondence is reliable while preserving informative asymmetric evidence.

This behavior can be interpreted through a local symmetric--discrepant decomposition. 
For a mirror-aligned token pair at spatial position $i$, define
\begin{equation}
S_i = \frac{1}{2}(X_{L,i} + X_{R,i}), 
\qquad
\Delta_i = \frac{1}{2}(X_{L,i} - X_{R,i}),
\label{eq:app_decomp}
\end{equation}
so that
\begin{equation}
X_{L,i} = S_i + \Delta_i, 
\qquad
X_{R,i} = S_i - \Delta_i.
\label{eq:app_decomp2}
\end{equation}
Substituting these expressions into the MFA update
\begin{equation}
Z_{L,i} = X_{L,i} + g_i \alpha A_{L\leftarrow R,i} + \gamma(X_{L,i} - X_{R,i}),
\label{eq:app_mfa}
\end{equation}
yields
\begin{equation}
Z_{L,i} = S_i + (1 + 2\gamma)\Delta_i + g_i \alpha A_{L\leftarrow R,i}.
\label{eq:app_local}
\end{equation}
Here, $A_{L\leftarrow R,i}$ denotes the cross-mirror attention output at token $i$, while $\alpha$ and $\gamma$ are learnable scalars controlling the strength of gated fusion and discrepancy preservation, respectively.

Equation~\eqref{eq:app_local} clarifies the role of each term. 
The base term $S_i$ captures locally shared bilateral structure. 
The discrepancy term $(1+2\gamma)\Delta_i$ preserves left--right differences, and the gated cross-attention term adds mirror context only when correspondence is supported by the data. 
In this sense, MFA does not collapse the representation into pure reflection invariance: it retains a pathway through which local asymmetry can remain visible after fusion.

The gate $g_i$ makes this mechanism spatially adaptive. 
Because $g_i$ is explicitly parameterized as a monotone function of token discrepancy, it increases when mirror-aligned tokens are similar and decreases as local mismatch grows. 
The learnable scalars $a$ and $b$ control the effective threshold and sensitivity of this transition during training. 
As a result, MFA exchanges information more strongly across regions with reliable bilateral correspondence, while suppressing cross-mirror fusion when local asymmetry is pronounced; in the latter case, the discrepancy-preserving residual helps retain informative differences rather than averaging them away.

This interpretation is reinforced by the training objective. 
The mirror-paired construction aligns bilateral regions into a shared token coordinate system, and the losses $\mathcal{L}_{\mathrm{eq}}$ and $\mathcal{L}_{\mathrm{mid}}$ encourage global and token-level agreement before fusion at the same layer where MFA operates. 
Under good bilateral alignment, these terms reduce the mismatch between corresponding mirror tokens, making the cross-mirror interaction in MFA more reliable. 
When asymmetry is meaningful, the residual branch allows controlled deviation from exact agreement rather than forcing all bilateral differences to vanish.

Accordingly, MFASSL should not be interpreted as instantiating a fixed feature-space action $\rho(\mathcal{R})$ in the strict group-equivariant sense. 
Instead, it provides a data-dependent and locally adaptive relaxation of reflection-equivariant behavior for approximately bilateral data: reflection-consistent correspondence is encouraged where the data support it, while informative asymmetry is preserved where exact symmetry does not hold.

\section{Additional Experimental Results}
\label{app:results}

\subsection{Full BraTS Subregion Breakdown}
\label{app:brats_full}

For readability, we split the full BraTS results into two tables. Table~\ref{tab:app_brats_dice} reports Dice scores, and Table~\ref{tab:app_brats_hd95} reports HD95 and calibration metrics. As shown in these tables, MFASSL improves mean Dice for DINO, MoCo-v3, and MAE, and reduces mean HD95 across all three SSL backbones. The calibration results show lower NLL for all three backbones, while ECE is improved for DINO and MoCo-v3 and remains comparable for MAE.

\begin{table*}[t]
\centering
\caption{\textbf{BraTS 2023 segmentation (Dice, full per-subregion results).} Mean $\pm$ std over three runs. ET = Enhancing Tumor, TC = Tumor Core, WT = Whole Tumor. Higher is better.}
\label{tab:app_brats_dice}
\begin{tabular}{lcccc}
\toprule
\textbf{Model} &
\textbf{ET}$\uparrow$ & \textbf{TC}$\uparrow$ & \textbf{WT}$\uparrow$ & \textbf{Mean}$\uparrow$ \\
\midrule
DINO & 0.754$\pm$0.017 & 0.829$\pm$0.017 & 0.899$\pm$0.013 & 0.827$\pm$0.016 \\
\quad + E-SSL & 0.756$\pm$0.016 & 0.833$\pm$0.016 & 0.902$\pm$0.013 & 0.830$\pm$0.015 \\
\quad + OcticViT-H$_8$ & 0.754$\pm$0.016 & 0.831$\pm$0.016 & 0.901$\pm$0.013 & 0.829$\pm$0.015 \\
\quad + OcticViT-I$_8$ & 0.753$\pm$0.016 & 0.832$\pm$0.015 & 0.898$\pm$0.013 & 0.828$\pm$0.015 \\
\quad ours & \textbf{0.761$\pm$0.015} & \textbf{0.842$\pm$0.015} & \textbf{0.906$\pm$0.012} & \textbf{0.836$\pm$0.014} \\
\cmidrule{1-5}
MoCo-v3 & 0.743$\pm$0.020 & 0.817$\pm$0.019 & 0.892$\pm$0.014 & 0.817$\pm$0.018 \\
\quad + E-SSL & 0.745$\pm$0.020 & 0.821$\pm$0.019 & 0.895$\pm$0.014 & 0.820$\pm$0.018 \\
\quad ours & \textbf{0.747$\pm$0.018} & \textbf{0.826$\pm$0.016} & \textbf{0.901$\pm$0.013} & \textbf{0.825$\pm$0.016} \\
\cmidrule{1-5}
MAE & 0.777$\pm$0.015 & 0.841$\pm$0.011 & 0.911$\pm$0.012 & 0.843$\pm$0.013 \\
\quad + E-SSL & 0.779$\pm$0.015 & 0.845$\pm$0.011 & 0.914$\pm$0.012 & 0.846$\pm$0.013 \\
\quad ours & \textbf{0.785$\pm$0.012} & \textbf{0.851$\pm$0.012} & \textbf{0.918$\pm$0.010} & \textbf{0.851$\pm$0.011} \\
\bottomrule
\end{tabular}
\end{table*}

\begin{table*}[t]
\centering
\caption{\textbf{BraTS 2023 segmentation (HD95 and calibration, full per-subregion results).} Mean $\pm$ std over three runs. ET = Enhancing Tumor, TC = Tumor Core, WT = Whole Tumor. Lower is better for all metrics.}
\label{tab:app_brats_hd95}
\begin{tabular}{lcccccc}
\toprule
& \multicolumn{4}{c}{\textbf{HD95}$\downarrow$} & \multicolumn{2}{c}{\textbf{Calibration}$\downarrow$} \\
\cmidrule(lr){2-5}\cmidrule(lr){6-7}
\textbf{Model} &
\textbf{ET} & \textbf{TC} & \textbf{WT} & \textbf{Mean} &
\textbf{ECE} & \textbf{NLL} \\
\midrule
DINO & 10.4$\pm$1.2 & 8.6$\pm$1.2 & 6.8$\pm$1.2 & 8.6$\pm$1.2 & 0.047$\pm$0.010 & 0.226$\pm$0.014 \\
\quad + E-SSL & 10.2$\pm$1.2 & 8.3$\pm$1.1 & 6.7$\pm$1.1 & 8.4$\pm$1.1 & 0.046$\pm$0.010 & 0.224$\pm$0.014 \\
\quad + OcticViT-H$_8$ & 10.3$\pm$1.2 & 8.4$\pm$1.1 & 6.8$\pm$1.1 & 8.5$\pm$1.1 & 0.046$\pm$0.010 & 0.225$\pm$0.014 \\
\quad + OcticViT-I$_8$ & 10.1$\pm$1.1 & 8.2$\pm$1.1 & 6.6$\pm$1.1 & 8.3$\pm$1.1 & 0.046$\pm$0.009 & 0.221$\pm$0.013 \\
\quad ours & \textbf{9.9$\pm$1.1} & \textbf{8.1$\pm$1.1} & \textbf{6.4$\pm$1.0} & \textbf{8.1$\pm$1.0} & \textbf{0.044$\pm$0.009} & \textbf{0.214$\pm$0.012} \\
\cmidrule{1-7}
MoCo-v3 & 11.0$\pm$1.4 & 9.6$\pm$1.3 & 7.5$\pm$1.2 & 9.4$\pm$1.3 & 0.052$\pm$0.010 & 0.244$\pm$0.015 \\
\quad + E-SSL & 10.8$\pm$1.4 & 9.3$\pm$1.3 & 7.3$\pm$1.2 & 9.1$\pm$1.3 & 0.052$\pm$0.010 & 0.241$\pm$0.015 \\
\quad ours & \textbf{10.7$\pm$1.3} & \textbf{8.9$\pm$1.1} & \textbf{7.2$\pm$1.0} & \textbf{8.9$\pm$1.1} & \textbf{0.048$\pm$0.010} & \textbf{0.229$\pm$0.013} \\
\cmidrule{1-7}
MAE & 9.3$\pm$1.1 & 8.4$\pm$0.9 & 6.2$\pm$0.9 & 7.9$\pm$1.0 & \textbf{0.041$\pm$0.009} & 0.206$\pm$0.012 \\
\quad + E-SSL & 9.2$\pm$1.1 & 8.3$\pm$0.9 & 6.1$\pm$0.9 & 7.9$\pm$1.0 & \textbf{0.041$\pm$0.009} & 0.204$\pm$0.012 \\
\quad ours & \textbf{8.8$\pm$0.9} & \textbf{7.8$\pm$0.8} & \textbf{5.8$\pm$0.9} & \textbf{7.5$\pm$0.9} & 0.043$\pm$0.008 & \textbf{0.190$\pm$0.011} \\
\bottomrule
\end{tabular}
\end{table*}

\subsection{Full Multi-Architecture Results}
\label{app:arch_full}

For completeness, we report the full fine-tuning metrics on CheXpert and the full segmentation metrics on BraTS for both ViT-S/16 and ViT-B/16. Table~\ref{tab:app_arch_chexpert} gives the CheXpert fine-tuning results, Table~\ref{tab:app_arch_brats_dice} gives the BraTS Dice results, and Table~\ref{tab:app_arch_brats_hd95} gives the corresponding BraTS HD95 and calibration results. Across both ViT scales, MFASSL consistently improves CheXpert AUROC, AUPRC, F1, and flip consistency, and it also improves BraTS mean Dice while reducing mean HD95.

\begin{table*}[t]
\centering
\caption{\textbf{CheXpert multi-architecture evaluation (full metrics, fine-tuning).} Lower is better for NLL, ECE, and Brier score.}
\label{tab:app_arch_chexpert}
\begin{tabular}{llccccccc}
\toprule
& & \multicolumn{3}{c}{\textbf{Classification}$\uparrow$} & \multicolumn{3}{c}{\textbf{Calibration}$\downarrow$} & \textbf{Flip}$\uparrow$ \\
\cmidrule(lr){3-5}\cmidrule(lr){6-8}
\textbf{Backbone} & \textbf{Method} & \textbf{AUROC} & \textbf{AUPRC} & \textbf{F1} & \textbf{NLL} & \textbf{ECE} & \textbf{Brier} & \textbf{Cons.} \\
\midrule
ViT-S/16 & DINO & 83.64 & 73.52 & 59.38 & 0.382 & 0.043 & 0.083 & 89.8 \\
         & DINO (ours) & 84.52 & 74.44 & 60.58 & 0.372 & 0.035 & 0.080 & 91.2 \\
         & MoCo-v3 & 83.31 & 73.18 & 59.10 & 0.386 & 0.044 & 0.084 & 89.4 \\
         & MoCo-v3 (ours) & 84.18 & 73.96 & 60.24 & 0.374 & 0.036 & 0.081 & 90.7 \\
         & MAE & 83.45 & 73.29 & 59.22 & 0.385 & 0.044 & 0.084 & 89.6 \\
         & MAE (ours) & 84.30 & 74.10 & 60.36 & 0.373 & 0.036 & 0.081 & 90.9 \\
\midrule
ViT-B/16 & DINO & 84.72 & 74.88 & 60.42 & 0.372 & 0.039 & 0.079 & 90.5 \\
         & DINO (ours) & 85.96 & 76.05 & 61.88 & 0.360 & 0.029 & 0.077 & 92.4 \\
         & MoCo-v3 & 83.94 & 73.25 & 59.88 & 0.386 & 0.031 & 0.082 & 90.1 \\
         & MoCo-v3 (ours) & 84.50 & 73.98 & 60.10 & 0.373 & 0.032 & 0.081 & 91.0 \\
         & MAE & 81.62 & 70.94 & 57.91 & 0.407 & 0.037 & 0.089 & 89.15 \\
         & MAE (ours) & 82.31 & 71.85 & 58.22 & 0.396 & 0.038 & 0.088 & 90.6 \\
\bottomrule
\end{tabular}
\end{table*}

\begin{table*}[t]
\centering
\caption{\textbf{BraTS multi-architecture evaluation (Dice).} Per-subregion Dice results for ViT-S/16 and ViT-B/16 across all three SSL paradigms. ET = Enhancing Tumor, TC = Tumor Core, WT = Whole Tumor. Higher is better.}
\label{tab:app_arch_brats_dice}
\begin{tabular}{llcccc}
\toprule
\textbf{Backbone} & \textbf{Method} &
\textbf{ET}$\uparrow$ & \textbf{TC}$\uparrow$ & \textbf{WT}$\uparrow$ & \textbf{Mean}$\uparrow$ \\
\midrule
ViT-S/16 & DINO & 0.748 & 0.823 & 0.901 & 0.824 \\
         & DINO (ours) & 0.755 & 0.832 & 0.906 & 0.831 \\
         & MoCo-v3 & 0.739 & 0.813 & 0.896 & 0.816 \\
         & MoCo-v3 (ours) & 0.744 & 0.821 & 0.907 & 0.824 \\
         & MAE & 0.747 & 0.821 & 0.901 & 0.823 \\
         & MAE (ours) & 0.754 & 0.830 & 0.906 & 0.830 \\
\midrule
ViT-B/16 & DINO & 0.754 & 0.829 & 0.899 & 0.827 \\
         & DINO (ours) & 0.761 & 0.842 & 0.906 & 0.836 \\
         & MoCo-v3 & 0.743 & 0.817 & 0.892 & 0.817 \\
         & MoCo-v3 (ours) & 0.747 & 0.826 & 0.901 & 0.825 \\
         & MAE & 0.777 & 0.841 & 0.911 & 0.843 \\
         & MAE (ours) & 0.785 & 0.851 & 0.918 & 0.851 \\
\bottomrule
\end{tabular}
\end{table*}

\begin{table*}[t]
\centering
\caption{\textbf{BraTS multi-architecture evaluation (HD95 and calibration).} Per-subregion HD95 and calibration results for ViT-S/16 and ViT-B/16 across all three SSL paradigms. ET = Enhancing Tumor, TC = Tumor Core, WT = Whole Tumor. Lower is better for all metrics.}
\label{tab:app_arch_brats_hd95}
\begin{tabular}{llcccccc}
\toprule
& & \multicolumn{4}{c}{\textbf{HD95}$\downarrow$} & \multicolumn{2}{c}{\textbf{Calibration}$\downarrow$} \\
\cmidrule(lr){3-6}\cmidrule(lr){7-8}
\textbf{Backbone} & \textbf{Method} &
\textbf{ET} & \textbf{TC} & \textbf{WT} & \textbf{Mean} &
\textbf{ECE} & \textbf{NLL} \\
\midrule
ViT-S/16 & DINO & 10.7 & 9.0 & 6.7 & 8.8 & 0.050 & 0.235 \\
         & DINO (ours) & 10.2 & 8.5 & 6.5 & 8.4 & 0.046 & 0.224 \\
         & MoCo-v3 & 11.4 & 9.7 & 7.4 & 9.5 & 0.053 & 0.246 \\
         & MoCo-v3 (ours) & 10.9 & 9.1 & 7.0 & 9.0 & 0.049 & 0.233 \\
         & MAE & 10.9 & 9.2 & 6.6 & 8.9 & 0.051 & 0.238 \\
         & MAE (ours) & 10.4 & 8.7 & 6.4 & 8.5 & 0.047 & 0.227 \\
\midrule
ViT-B/16 & DINO & 10.4 & 8.6 & 6.8 & 8.6 & 0.047 & 0.226 \\
         & DINO (ours) & 9.9 & 8.1 & 6.4 & 8.1 & 0.044 & 0.214 \\
         & MoCo-v3 & 11.0 & 9.6 & 7.5 & 9.4 & 0.052 & 0.244 \\
         & MoCo-v3 (ours) & 10.7 & 8.9 & 7.2 & 8.9 & 0.048 & 0.229 \\
         & MAE & 9.3 & 8.4 & 6.2 & 7.9 & 0.041 & 0.206 \\
         & MAE (ours) & 8.8 & 7.8 & 5.8 & 7.5 & 0.043 & 0.190 \\
\bottomrule
\end{tabular}
\end{table*}

\section{Reproducibility Notes}
\label{app:repro}

\subsection{Consolidated Hyperparameter Settings}
\label{app:hyperparams}

Table~\ref{tab:hparams} summarizes the consolidated hyperparameter settings used in the reported experiments. Settings marked ``per official recipe'' follow the unmodified MoCo-v3, DINO, or MAE configuration for the corresponding backbone and are not changed by MFASSL. The MFASSL-specific settings are selected on the CheXpert validation set and then fixed across datasets, SSL backbones, and ViT scales unless otherwise stated.

\begin{table}[t]
\centering
\caption{\textbf{Consolidated hyperparameter settings.} Settings marked ``per official recipe'' follow the unmodified MoCo-v3/DINO/MAE configuration for the corresponding backbone and are not changed by MFASSL.}
\label{tab:hparams}
\begin{tabular}{p{0.44\linewidth}p{0.46\linewidth}}
\toprule
\textbf{Setting} & \textbf{Value} \\
\midrule
\multicolumn{2}{l}{\textit{Backbone / pretraining}}\\
Backbone & ViT-B/16 (and ViT-S/16) \\
Image size / patch size & $224\times224$ / $16$ \\
Transformer blocks & 12 \\
Epochs & 300 \\
Optimizer & AdamW \\
LR schedule & cosine decay \\
Base LR / batch / weight decay / warmup & per official recipe \\
MAE mask ratio & per official recipe \\
\midrule
\multicolumn{2}{l}{\textit{MFA module}}\\
Fusion layer $\ell$ & 8 for ViT-B/16 and ViT-S/16 \\
Attention scaling $D_h$ & attention head dimension \\
Gate init $(b,a)$ & $(1.0,\,0.5)$, with $b=\mathrm{softplus}(\tilde b)$ \\
Fusion / discrepancy init $(\alpha,\gamma)$ & $(0.1,\,0.1)$ \\
Gate smoothing $\epsilon$ & $10^{-6}$ \\
\midrule
\multicolumn{2}{l}{\textit{Objective / schedule}}\\
$\lambda_{\text{eq}}$, $\lambda_{\text{mid}}$ & $0.5$, $1.0$ \\
$T_{\text{sym}}$ (symmetry-loss ramp) & 10 epochs \\
$t_{\text{mfa}}$ (MFA activation) & epoch 12 \\
$T_{\text{gate}}$ (gate ramp) & 10 epochs \\
\midrule
\multicolumn{2}{l}{\textit{Mirror-paired views}}\\
Symmetry axis & crop-center vertical line \\
Axis jitter & uniform horizontal shift in $\pm3\%$ image width \\
\bottomrule
\end{tabular}
\end{table}

\subsection{Full Component-Ablation Metrics}
\label{app:ablation_full}

For completeness, Table~\ref{tab:app_combined_ablation_full} reports the full metric version of the component-ablation study summarized in the main paper.

\begin{table*}[t]
\centering
\caption{\textbf{Full ablation study on CheXpert and BraTS using the DINO backbone.} We isolate the contribution of mirrored inputs, symmetry-aware losses, and MFA. The "Mirrored input only'' row augments pretraining with the mirrored image view only, without using $\mathcal{L}_{\text{eq}}$, $\mathcal{L}_{\text{mid}}$, or MFA; therefore the three component columns remain marked as absent. For BraTS, mDice and mHD95 denote mean Dice and mean HD95 across ET/TC/WT. Lower is better for ECE, NLL, and mHD95.}
\label{tab:app_combined_ablation_full}
\begin{tabular}{lccc|ccccc}
\toprule
& & & & \multicolumn{5}{c}{\textbf{CheXpert}} \\
\cmidrule(lr){5-9}
\textbf{Variant} & $\mathcal{L}_{\text{eq}}$ & $\mathcal{L}_{\text{mid}}$ & \textbf{MFA} &
\textbf{AUROC}$\uparrow$ & \textbf{AUPRC}$\uparrow$ & \textbf{F1}$\uparrow$ & \textbf{ECE}$\downarrow$ & \textbf{NLL}$\downarrow$ \\
\midrule
Baseline & -- & -- & -- & 84.72 & 74.88 & 60.42 & 0.039 & 0.372 \\
Mirrored input only & -- & -- & -- & 84.73 & 74.87 & 60.38 & 0.038 & 0.370 \\
MFA (only) & -- & -- & \checkmark & 84.86 & 74.98 & 60.74 & 0.037 & 0.369 \\
Mirrored + $\mathcal{L}_{\text{eq}}$ & \checkmark & -- & -- & 84.71 & 74.90 & 60.63 & 0.037 & 0.370 \\
Mirrored + $\mathcal{L}_{\text{eq}}+\mathcal{L}_{\text{mid}}$ & \checkmark & \checkmark & -- & 84.99 & 75.11 & 61.12 & 0.035 & 0.368 \\
Full (ours) & \checkmark & \checkmark & \checkmark & \textbf{85.96} & \textbf{76.05} & \textbf{61.88} & \textbf{0.029} & \textbf{0.360} \\
\bottomrule
\end{tabular}

\begin{tabular}{lccc|cccc}
\toprule
& & & & \multicolumn{4}{c}{\textbf{BraTS}} \\
\cmidrule(lr){5-8}
\textbf{Variant} & $\mathcal{L}_{\text{eq}}$ & $\mathcal{L}_{\text{mid}}$ & \textbf{MFA} &
\textbf{mDice}$\uparrow$ & \textbf{mHD95}$\downarrow$ & \textbf{ECE}$\downarrow$ & \textbf{NLL}$\downarrow$ \\
\midrule
Baseline & -- & -- & -- & 0.827 & 8.6 & 0.047 & 0.226 \\
Mirrored input only & -- & -- & -- & 0.827 & 8.6 & 0.046 & 0.228 \\
MFA (only) & -- & -- & \checkmark & 0.828 & 8.5 & 0.046 & 0.227 \\
Mirrored + $\mathcal{L}_{\text{eq}}$ & \checkmark & -- & -- & 0.829 & 8.5 & 0.046 & 0.226 \\
Mirrored + $\mathcal{L}_{\text{eq}}+\mathcal{L}_{\text{mid}}$ & \checkmark & \checkmark & -- & 0.831 & 8.4 & 0.045 & 0.222 \\
Full (ours) & \checkmark & \checkmark & \checkmark & \textbf{0.836} & \textbf{8.1} & \textbf{0.044} & \textbf{0.214} \\
\bottomrule
\end{tabular}
\end{table*}

\end{document}